\documentclass[letterpaper, 10 pt, conference]{ieeeconf}  % Comment this line out if you need a4paper

\IEEEoverridecommandlockouts                              % This command is only needed if 
\usepackage{subcaption}
\usepackage{graphicx}
\usepackage{multirow}

\usepackage[utf8]{inputenc}
\usepackage[ruled, lined, linesnumbered, commentsnumbered, longend]{algorithm2e}

\usepackage{verbatim}

\usepackage{pgfplots}
\pgfplotsset{compat = 1.8}
\colorlet{xcolorA}{red!80!black}
\colorlet{xcolorB}{orange}
\colorlet{ycolorA}{blue}
\colorlet{ycolorB}{blue!40!white}

\title{\LARGE \bf
Accelerating Probabilistic Volumetric Mapping using Ray-Tracing Graphics Hardware}

\author{Heajung Min, Kyung Min Han and Young J. Kim% <-this % stops a space
\thanks{The authors are with the department of computer science and engineering at Ewha womans university in Korea  {\tt\small hjmin@ewhain.net,  \{hankm|kimy\}@ewha.ac.kr}}
}

\begin{document}

\maketitle
\thispagestyle{empty}
\pagestyle{empty}

%%%%%%%%%%%%%%%%%%%%%%%%%%%%%%%%%%%%%%%%%%%%%%%%%%%%%%%%%%%%%%%%%%%%%%%%%%%%%%%%
\begin{abstract}
Probabilistic volumetric mapping (PVM) represents a 3D environmental map for an autonomous robotic navigational task. A popular implementation such as Octomap is widely used in the robotics community for such a purpose. The Octomap relies on octree to represent a PVM and its main bottleneck lies in massive ray-shooting to determine the occupancy of the underlying volumetric voxel grids.

In this paper, we propose GPU-based ray shooting to drastically improve the ray shooting performance in Octomap. Our main idea is based on the use of recent ray-tracing RTX GPU, mainly designed for real-time photo-realistic computer graphics and the accompanying graphics API, known as DXR. Our ray-shooting first maps leaf-level voxels in the given octree to a set of axis-aligned bounding boxes (AABBs) and employ massively parallel ray shooting on them using GPUs to find free and occupied voxels. These are fed back into CPU to update the voxel occupancy and restructure the octree. 
In our experiments, we have observed more than three-orders-of-magnitude performance improvement in terms of ray shooting using ray-tracing RTX GPU over a state-of-the-art Octomap CPU implementation, where the benchmarking environments consist of more than 77K points and 25K$\sim$34K voxel grids.
\end{abstract}

%%%%%%%%%%%%%%%%%%%%%%%%%%%%%%%%%%%%%%%%%%%%%%%%%%%%%%%%%%%%%%%%%%%%%%%%%%%%%%%%
\section{INTRODUCTION}

%1. Talk about the importance of probabilistic volumetric mapping in robotics (SLAM, NAVIGATION, ...)
%2. Talk about why representing the map using Octree is necessary and how popular it is (Octomap, ...)
%3. Talk about a general strategy of how the probabilistic vol. mapping is created/updated based on ray shooting on Octree (insert scan?)
%4. Talk about what the bottleneck in 3 is and why it is slow.

%1. Talk about the importance of probabilistic volumetric mapping in robotics (SLAM, NAVIGATION, ...)

3D mapping is an essential component for autonomous navigational tasks since the accuracy of 3D mapping significantly affects estimating the surroundings where the robot is deployed. The reconstructed map subsequently impacts the quality of trajectories predicted by a motion planner whose goal is correctly guiding the robot in the given environment.
Probabilistic volumetric mapping (PVM) is a popular strategy for representing such 3D maps for two main reasons. First, in PVM, a map is represented by a set of voxels that adaptively subdivides a 3D space depending on the occupancy of the space. As such, the adaptive voxel representation reduces a substantial amount of memory required to represent the 3D environment. Second, the occupancy of a voxel is represented probabilistically \cite{ThrBur05}, empowering the map to cope with uncertainties including sensor noise and dynamic scenarios.    

%2. Talk about why representing the map using Octree is necessary and how popular it is (Octomap, ...)
As a choice for volumetric reconstruction, an octree is a \emph{de facto} standard thanks to its practical benefits on both computational and memory efficiencies. An octree is a spatial subdivision that adequately represents the hierarchical nature of voxels, in particular where the time complexity for traversal is maintained in the logarithmic scale.   

%3. Talk about a general strategy of how the probabilistic vol. mapping is created/updated based on ray shooting on Octree (insert scan?)
Octomap \cite{HorBur13} is a state-of-the-art implementation for PVM using the octree representation that consists of the following steps to build a PVM:
\begin{enumerate}
    \item Scan and generate a point cloud for the environment.
    \item Shoot rays toward each point in the point cloud.
    \item Find and identify free or occupied voxels in the observed space.
    \item Update the octree with updated occupancy.
\end{enumerate}
%4. Talk about what the bottleneck in 3 is and why it is slow.
Often, the second and third steps, i.e., the ray-shooting step, are the most time-consuming operation, especially when the input point cloud is large or the ray-length becomes long. Such cases potentially limit the PVM module to maintain the map at a coarse level, 
%For instance, if the input point cloud is obtained from dense stereo correspondences, the size of the point cloud would be proportional to the resolution of the camera (e.g., 307200 points for $640\times480$ sized image pairs.) 
and the situation will get worse if a higher-resolution sensor is used to generate a point cloud.
Furthermore, the slow ray-shooting process could hinder the robot from conducting online navigation.

%The issue with the CPU-based ray shooting is that it takes a lot of ray shooting time. We measured that the ray tracing time occupied 90\% of the total time to update the occupancy state and reconstruct the octree including ray tracing for voxels. So, we want to reduce the time to finally construct a 3D volume map by improving the performance by performing ray tracing using GPU acceleration.

%Talk about the main idea of paper including octomap-rtx mapping and ray shooting
{\bf Main Results:}
In this paper, we perform GPU-based ray shooting to drastically improve the ray shooting performance in Octomap, which is the main bottleneck in octree-based PVM. 
%A ray traces a voxelized 3D environment from the camera position to the 3D coordinate of the point cloud representing the surface of an obstacle from a distance sensor. 
%CPU-based ray shooting searches voxels intersecting from the ray origin on the lowest-level voxel grid sequentially and these voxels are stored as the free space and the voxel stored at the end point from the ray become occupied.
%On the other hand, we perform ray tracing in the 3d environment represented by octree. 
The main idea of our work is mapping leaf-level voxels in the given octree to a set of axis-aligned bounding boxes (AABBs) and employing massively parallel ray shooting on them using GPUs. 
%That is, the voxel of the octree is configured in the form of Axis-Aligned Bounding Box(AABB) used by the GPU ray tracer, and AABB that intersects the ray is quickly found through GPU acceleration. 
The intersected AABBs are further subdivided into the finest resolution corresponding to free or occupied voxels. After that, voxel occupancy and octree restructuring are followed on the CPU after the voxels are readback from the GPU. 
%Talk about experimmental results
In our experiments, we have observed more than three-orders-of-magnitude performance improvement in terms of ray shooting using ray-tracing RTX GPU over a state-of-the-art Octomap CPU implementation, where the benchmarking environments consist of 77K points and 25K$\sim$34K adaptive voxel grids.

%Octomap is an open source in the field of Robotics as an octree-based occupancy map. It constructs a 3d environment scanned with a point cloud using a range sensor into voxels based on an octree and represents it as a map using probabilistic occupancy estimation. For this, Octomap performs occupancy inquire for each voxel, which distinguishes whether the voxel state is an occupied voxel or a free voxel at a specific view using ray tracing according to each point cloud being located. 

%In other words, the voxelized 3D environment is traced by generating a ray starting from the camera position. The starting point of each ray is set by the camera position, and the direction and length of the ray are calculated using the end point of the ray defined for each ray. The end point refers to the 3D coordinate of the point cloud representing the surface of an obstacle from a distance sensor. The voxel containing this end point becomes an occupied space. 

%When ray tracing starts, voxels intersecting from the starting point of the ray are sequentially searched and these voxels are stored as free space. Ray tracing is performed only within the length of the ray, and tracing is stopped for lengths beyond this.

%As a result, the voxel stored at the end point from the ray becomes occupied, and the voxel intersecting the ray at the closer position becomes free. After updating the occupancy state of these voxels using probabilistic occupancy estimation, the octree is reconstructed to represent the 3D volume map.

\section{RELATED WORK}

\subsection{3D Volumetric Mapping}

The idea of subdividing a 2D planar floor into uniform grid cells, namely occupancy grid map (OGM), dates back to mid-80s \cite{MorElf85}. Later, the grid map idea was expanded to reconstruct a 3D map where a volumetric space is subdivided into equal-sized volumetric grids, i.e., voxels \cite{RotJai89,Mor96}. However, representing a volumetric space with uniform grids inevitably causes memory problems, preventing practical applications demanded for large scale and long-term operations. A well-known strategy to mitigate this problem is employing an octree data structure \cite{HorBur13, SchZel14, VesLeu18}. 

The voxel-based sampling assumes that the occupancy of each voxel is independent of its neighboring voxels. While this assumption is useful for simplifying the OGM, the resulting map suffers from inaccuracy to some extent. Gaussian Process (GP)-based methods \cite{CalRam11,VasBla09}, on the other hand, consider a continuous spatial domain rather than discretized grid cells. As a result, GP allows maps to estimate unknown terrains. Besides, the map can represent itself with various resolutions. Unfortunately, the drawback of this method is a large memory requirement and a cubic time complexity in terms of the number of cells. Subsequently, there have been researches to ameliorate the performance issue of GP in recent years \cite{KimKim12,WanEng16}.  

When the application of the map is limited to particular purposes such as the legged robot's motion planning, a 2.5D height map \cite{FanHut18,BelSkr12} is an efficient way to reconstruct the local surroundings of the robot. %In this case, a 3D space is often rendered by only a 2D grid map where each grid cell contains height information of the cell position.     
Normal Distributions Transform (NDT) is another approach to discretize a 3D volumetric space. In contrast to voxel-based sampling, a cell in NDT contains multiple points to form a local Gaussian distribution. For this reason, NDT is considered as a piece-wise continuous representation of a space, where the number of grid cells is much smaller than that of voxel grid maps. NDT was first proposed for 2D scan registration purposes \cite{BibStr03}. Later, the idea was further developed to 3D scan registration methods \cite{MagDuc07,StoLil12}. \cite{SaaLil13icra_ndtom} proposed to augment occupancy probability to NDT, which was followed by a real-time version \cite{SchZel18}.

\subsection{GPU-based Octree Construction}\label{sec:gpurecon}

The octree structure is predominantly used in computer graphics for various tasks including distance
field generation, rendering, modeling, simulation and model reconstruction~\cite{HorBur13,hoetzlein2016gvdb,liu2014exact}.
% presented a method to compute ADFs on GPUs. 
\cite{zhou2010data} proposes a GPU-based octree construction for reconstructing surfaces on the GPU. 
% octree update
For a large-scale volumetric scene, full or out-of-core style octree update to GPU was studied~\cite{museth2013vdb,hadwiger2012interactive}.
Octree can be adjusted dynamically in real-time in GPU~\cite{crassin2011interactive} as well as one-time construction or full reconstruction~\cite{crassin2009gigavoxels,debry2002painting}.
\cite{gobbetti2008single} studied streaming subtree data through CPU-GPU data transfer in a view-dependent manner with connectivity information. Recently, \cite{hoetzlein2016gvdb} supports dynamic topological updates on GPU.

% octree construction
To reduce the cost of searching neighbors during ray traversals on Octree, \cite{gobbetti2008single} has reduced the number of neighbors down to six per cell by pointing the parents of neighbors. \cite{kim2018dynamic} used three precomputed neighbors per a cell that enable stackless ray casting and dynamic updating of Octree on a GPU.
A sparse voxel octree (SVO) showed both high-quality rendering and efficient ray traversal of shallow tree topology for the static scene~\cite{laine2010efficient}. OpenVDB~\cite{museth2013vdb} used SVO data structures and was implemented in GPU~\cite{hoetzlein2016gvdb}, which enables efficient neighbor access using GPU-based ray casting for dynamic scenes.

\subsection{GPU-based Ray Tracing}
Ray tracing is a graphical technique to render a realistic scene using the physical properties of light. From a given camera viewpoint, many rays are generated and shot toward the virtual 3D scene and each ray-path is traced to determine the corresponding pixel color of the screen~\cite{Shirley}. 

% shortly
Graphics hardware-based ray tracing has been studied to accelerate each stage of the ray-tracing pipeline including constructing acceleration structure, ray generation, traversing acceleration structure, and ray-triangle intersection.
The growing need for rendering dynamic scenes demanded online construction and update of the acceleration structure, such as kd-tree~\cite{woop5programmable} or BVH~\cite{doyle2013hardware}. ~\cite{nah2014raycore} introduced dedicated graphics hardware for ray generation. 
The ray traversal is the bottleneck of the ray tracing pipeline. ~\cite{woop2007programmable} explored the space of ray traversal on kd-trees, and ~\cite{lee2013sgrt} proposed a BVH traversal hardware using three parallel pipelines. After traversal, the process of identifying the closest intersection point through the ray-triangle hit test has been studied ~\cite{lee2013sgrt,nah2014raycore}.

In recent years, the RTX GPU platform for accelerating ray tracing has been introduced, which accelerates the BVH construction and traversal, and proposes a new rendering pipeline with a machine learning-based denoising technique to enable real-time ray tracing rendering~\cite{RTX,dxrTutorial}. 
% gpgpu?
Moreover, RTX can be used as general-purpose computing (GPGPU) for non-rendering tasks, such as sampling for simulation~\cite{gribble2019multi}, point location in tetrahedral meshes~\cite{wald2019rtx}, or Monte Carlo particle transport~\cite{salmon2019exploiting}.

%\subsection{Selecting a Template (Heading 2)}

%First, confirm that you have the correct template for your paper size. This template has been tailored for output on the US-letter paper size. 

%\subsection{Maintaining the Integrity of the Specifications}

%The template is used to format your paper and style the text.

%\section{OVERVIEW}

%The template is used to format your paper and style the text.

\section{Fast and Massively-Parallel Ray Shooting on RTX}

%\subsection{...}

%\subsection{RTX Pipeline}
Since ray-tracing is a computationally intensive technique, dedicated hardware support is highly beneficial. Timely, NVIDIA introduced and started to mass-produce ray tracing hardware, known as RTX. 
In CPU-based ray tracing, for each ray, thousands of serial CPU instructions need to be executed to check the ray-surface intersection using the bounding volume hierarchy (BVH) representing the scene geometry, often composed of many triangles.
Many RT cores installed in the RTX graphics hardware consist of two units, one responsible for BVH traversal and the other for ray-triangle or ray-AABB intersection test~\cite{ARCHITECTURE}.
%As a result, the number of processed rays per second has increased by about almost 35 times compared to the CPU~\cite{RtCore}.

In order to harness the power of RTX GPU, an application programming interface (API) such as DirectX's DXR has been also introduced.
In DXR, scene geometry is divided into bottom/top level acceleration structures, which is essentially a BVH. In the bottom level acceleration structure, geometric primitives such as triangles or AABBs are included, and in the top level acceleration structure, transformation matrix and rendering material are included.  

Also, shader programs in the RTX pipeline are executed in a massively-parallel fashion for all rays. Specifically, first, after a BVH is built, rays are generated in {\em the ray generation shader}, traverse the BVH, and check intersection for hitting bounding volumes. Then, for the hit group, 
{\em the intersection shader} finds an intersection of the ray and the geometry, and {\em the closest hit shader} calculates shading information at the hit point. If the ray does not intersect any geometry, {\em the miss shader} would be executed~\cite{RTX}.
%Ray generation, miss shader are not part of hit groups because they aren’t involved directly with geometry. 
In our application, we do not use the closest hit shader as no shading information is needed.

\section{GPU-ACCELERATED RAY SHOOTING}

\subsection{Octree Representation of PVM}
%describe Octree representation in Octomap
The octree is a hierarchical data structure created by recursively subdividing three-dimensional space. Each octree node represents a small volumetric grid called a voxel and has eight sub-voxels as child nodes. The octree is also useful for labeling its subspace with different occupancy information using voxels. 

A well-known PVM such as Octomap~\cite{HorBur13} relies on octree to map the environment into a set of voxels representing a {\em free} or a {\em occupied} state (obstacle space). Specifically, Octomap builds a 3D map by repeatedly relocating a proximity sensor at different positions and carrying out the following steps in turn: 
\begin{enumerate}
    \item A sensor observation is obtained in the form of a point cloud.
    \item The sensor origin $\mathbf{o}$ and each point $\mathbf{p}$ in the point cloud constitute an individual ray $\overrightarrow{\mathbf{op}}$ with a fixed length $|\overrightarrow{\mathbf{op}}|$.
    \item For all rays, a ray-shooting procedure is performed against the voxelized space using DDA~\cite{AmeWoo87} to discover free and occupied voxels. These labeled voxels correspond to the finest resolution of the octree. 
    \item The discovered voxels are merged into the octree and restructured depending on the cells' probabilistic occupancies.
\end{enumerate}
Note that the second and the third step take up about 90\% of processing time for each scan and is the main bottleneck. The main goal of our work is to drastically improve it using massively parallel ray shooting on GPU.
 %Here, each node's occupancy is updated based on the log-odd formula, where additions replace multiply operations. Hence, prompt occupancy updates are guaranteed. Besides, the log-odd occupancy estimates are clamped by min and max thresholds in order to facilitate fast adaptation to a dynamic environment. Once the nodes' occupancies are updated, a pruning procedure eliminates eight octants with the same occupancy state. Therefore, a compact map is maintained over the map building task.

\subsection{Mapping an Octree on CPU to AABBs on GPU} 

%In this work, we find voxels and update their occupancy information through massive GPU-based ray shooting.
%Specifically, we shoot many rays toward occupied voxels to find voxels corresponding to free space. 
%Space that is not initialized to occupied or free implicitly model unknown voxel.
%Accordingly, it is possible to plan a collision-free path with the spatial configuration of such an octree node.

In order to leverage GPU-accelerated ray shooting for an octree, the voxel elements of the octree must be converted to geometric primitives that the ray-tracing GPU such as the RTX can process, which should be a set of triangles or AABBs in case of RTX. In our problem, we opt for AABBs as target primitives as their geometries are close to the shape of a voxel.

\begin{figure}[htb]
     \centering
     \begin{subfigure}[b]{0.35\linewidth}
         \centering
         \includegraphics[width=\linewidth]{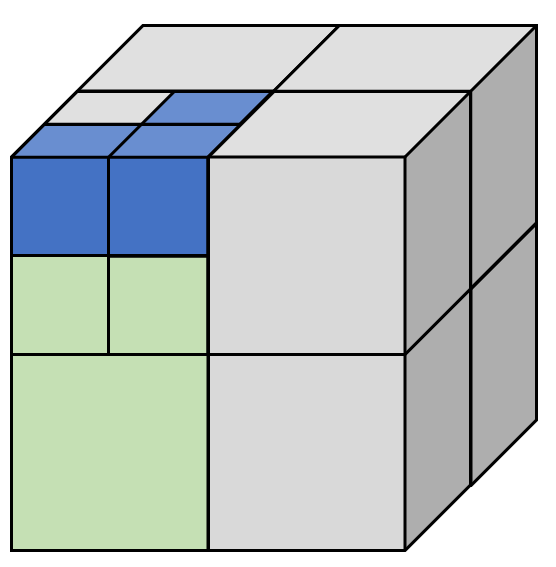}
         \caption{\label{fig:aabb_0}
            Spatial Subdivision}
     \end{subfigure}
     \hfill
     
     \begin{subfigure}[b]{0.9\linewidth}
         \centering
         \includegraphics[width=\linewidth]{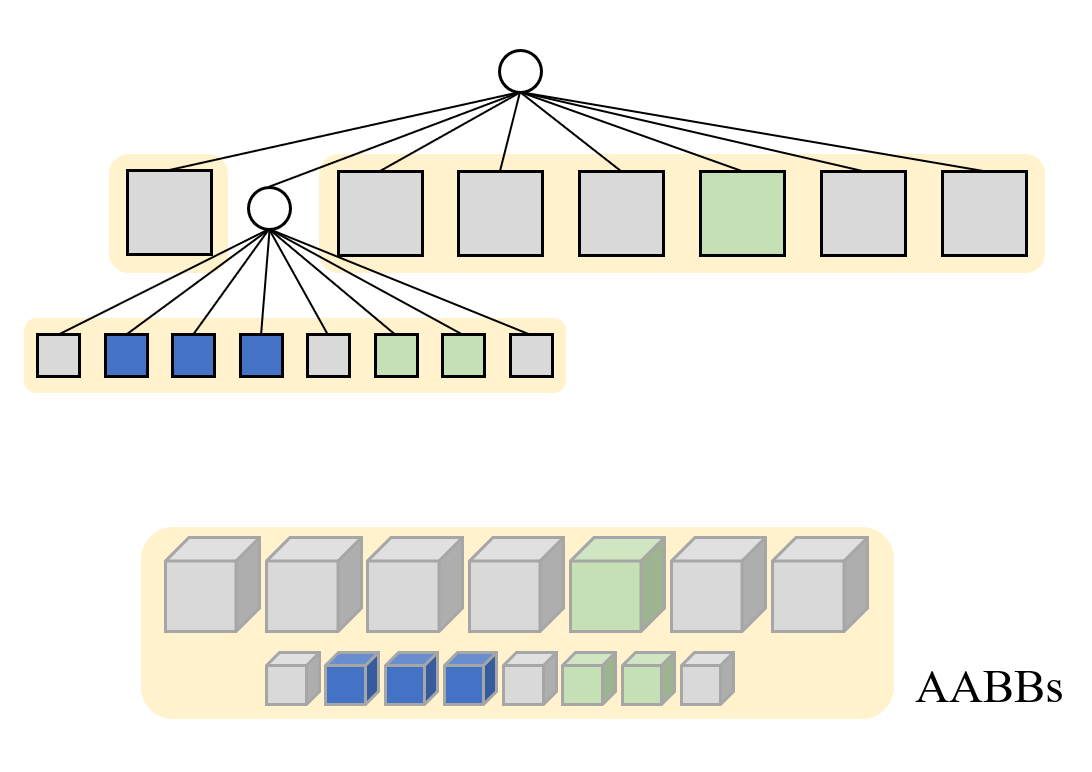}
         \caption{\label{fig:aabb_1}
                Octree and the corresponding AABBs}
     \end{subfigure}
     \hfill
     
    \caption{\label{fig:octree}
        Mapping from an octree in CPU to a set of AABBs in GPU for GPU-based ray shooting. In the figures, blue, green, and gray boxes indicate occupied, free, unknown voxel space, respectively.
        }
\end{figure}

Specifically, we convert all leaf-level voxels in an octree, with occupied (blue), free (green), and unknown (gray) labels, to individual AABBs. 
As illustrated in Fig.~\ref{fig:octree}, a leaf-level voxel is mapped to an AABB by matching their geometries including the position and size. 
Note that we also consider voxels with an unknown label during the conversion so that the AABBs need to fill the entire 3D workspace since a random ray can pass anywhere in the space and the traversed subspace needs to be labeled afterward.

After all leaf-level voxels are converted to a set of AABBs, they are uploaded to the GPU, and the RTX GPU builds a bounding volume hierarchy (BVH) of AABBs.
Fig.~\ref{fig:octree} illustrates (a) spatial subdivision of space with different labels to represent occupied, free, or unknown voxel space, (b) a corresponding octree representation on CPU with leaf-level nodes highlighted in yellow and a set of AABBs on GPU that correspond to the leaf-level voxel nodes on CPU.

%\begin{figure}[h]
%     \centering
%     \begin{subfigure}[b]{0.75\linewidth}
%         \centering
%         \includegraphics[width=\linewidth]{bvh_0.PNG}
%         \caption{\label{fig:bvh_0}
%            }
%     \end{subfigure}
%     \hfill
%     
%     \begin{subfigure}[b]{0.6\linewidth}
%         \centering
%         \includegraphics[width=\linewidth]{bvh_1.png}
%         \caption{\label{fig:bvh_1}
%                }
%     \end{subfigure}
%     \hfill
%     
%    \caption{\label{fig:bvh}
%        (a) Bounding volume hierarchy setup with AABBs (b) Ray and AABB intersection.
%        }
%\end{figure}

%After uploading AABBs to the GPU, BVH is built as a tree structure with many levels. Fig.~\ref{fig:bvh_0} illustrates each node is encompassed by a bounding box that bounds all of its descendant nodes. Each leaf node includes the AABB respectively.

\subsection{Massively-Parallel Ray Shooting}

%For efficient probabilistic volume mapping, we apply GPU-based ray shooting to update the occupancy of a volume presented in the octree. We serve occupied and free voxels explicitly that are generated in the space between the sensor and the obstacle. 

Once the BVH of AABB is computed on GPUs, we set up multiple rays in the ray generation shader and shoot them from the sensor origin to the environment obstacles, obtained as a point cloud by the sensor, to find occupied or free voxels of space.
The direction of each ray is defined by the sensor origin (ray start) and the position of each point (ray end) in the cloud. 
Therefore, the number of rays is proportional to the number of points in the cloud, and the rays are shot in parallel. 
We set each ray's length equal to the distance from the sensor origin to the point cloud. This way, the ray end always corresponds to an occupied voxel and the interior of the ray to free voxels.

\begin{figure}[htb]
\centering
 \begin{subfigure}[b]{0.4\linewidth}
        %  \centering
         \includegraphics[width=\linewidth]{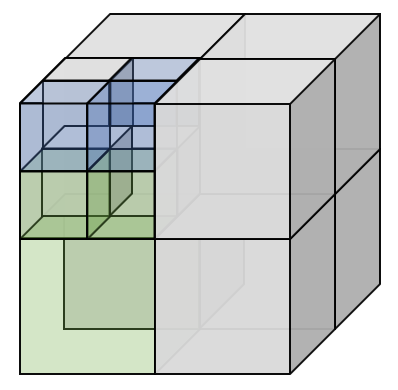}
        \caption{Mapped AABBs}
     \end{subfigure}
     \begin{subfigure}[b]{0.4\linewidth}
        %  \centering
         \includegraphics[width=\linewidth]{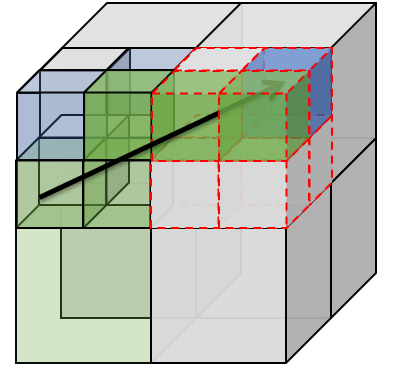}
         \caption{Subdivision}
     \end{subfigure}
    
    \caption{\label{fig:octree_ray_aabb}
        Leaf-level voxels are mapped to a set of AABBs and subdivided to the finest resolution for occupancy labeling.
        }
\end{figure}

\begin{figure}[htb]
     \centering
      \begin{subfigure}[b]{0.44\linewidth}
        %  \centering
         \includegraphics[width=\linewidth]{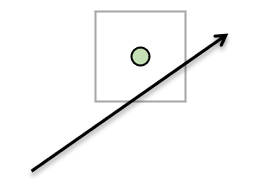}
        \caption{Intersecting a ray}
     \end{subfigure}
     \begin{subfigure}[b]{0.37\linewidth}
        %  \centering
         \includegraphics[width=\linewidth]{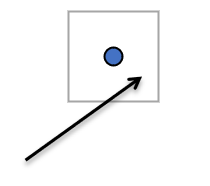}
         \caption{Including the ray end}
     \end{subfigure}
     %\hfill

     %\hfill
     \begin{subfigure}[b]{0.5\linewidth}
         \centering
         \includegraphics[width=\linewidth]{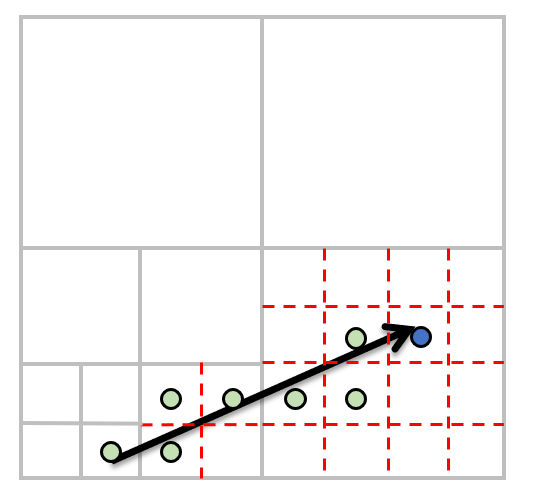}
        \caption{Intermediate voxel}
     \end{subfigure}
     %\hfill
    \caption{Different cases of ray-AABB intersection. (a), (b) a ray hitting finest-resolution voxels, (c) a ray hitting a intermediate voxel that is not the finest resolution. (a) is labeled as free (green) and (b) is as occupied (blue). (c) is subdivided into finest resolution and labeled accordingly.}\label{fig:ray_aabb}
\end{figure}

\begin{algorithm}[h]
\caption{GPU-based Ray Shooting}\label{alg:GPU_rayTracing}
\SetAlgoLined
% \DontPrintSemicolon
\SetKwData{Left}{left}\SetKwData{This}{this}\SetKwData{Up}{up}
\SetKwFunction{Union}{Union}\SetKwFunction{FindCompress}{FindCompress}
\SetKwBlock{DoParallel}{do in parallel}{end}
\SetKwInOut{Input}{Input}\SetKwInOut{Output}{Output}

\Input{sensorOrigin, rayDirections, a set of AABBs}
\Output{Intersected voxels with the rays}
\BlankLine

Build the BVH of AABBs in parallel\;
\DoParallel{
\tcc{using Ray Generation shader}
Setup a ray\;
\While{The ray hits AABBs within the ray extent during BVH traversal}{
    \tcc{using Intersection shader}
    \If{The AABB size is greater than finest resolution voxels}{
    Subdivide the AABB to a set of finest resolution voxels\;}
    \ForAll{Newly generated finest resolution voxels}{
          \uIf{voxel includes the ray's end point}{
          Set it as occupied\;
          }
          \uElseIf{ray passed the voxel}{
          Set it as free\;
          }
      }
    }
}
\end{algorithm}

%Then, the ray tracer starts looking for AABB that intersects with the ray. 
%Fig.~\ref{fig:bvh_1} illustrates the process that starts by testing the ray against the root node bounding box of BVH and working down the tree of descendant nodes to test which successively smaller bounding boxes are intersected by the ray. 
%It is guaranteed to find any AABB intersected within the ray extent, regardless of the order of distance from the ray %origin. Intersecting AABBs outside this ray interval is not permitted.

In our current implementation, we rely on GPUs to perform ray shooting, but still use CPU for octree and occupancy update, which is an effective strategy as the update step is not a dominant part of the entire pipeline. In particular, we use Octomap for the update part, and Octomap expects voxels of the finest resolution in the octree as a result of ray shooting.
In order to meet this interfacing requirement, in our GPU-based ray shooting, if the size of an AABB intersected with a ray is greater than that of a finest-resolution voxel, we subdivide the subspace that the ray traverses to a set of sub-voxels in the finest resolution (typically 16) using DDA~\cite{AmeWoo87}.  
These voxels are labeled as free except that the voxel containing the end point of the ray is labeled as occupied.
Fig.~\ref{fig:octree_ray_aabb} illustrates this situation and Fig.~\ref{fig:ray_aabb} illustrates different cases of ray-AABB intersection. When ray-AABB intersection occurs, determining the occupancy of the voxel is performed in the intersection shader.
If the intersection of ray and AABB no longer occurs, the miss shader is executed. Since we do not need shading in our work, we execute the miss shader to simply terminate the ray shooting.\newline
\indent After a set of voxels hit by rays are found, their occupancy is updated, and the octree nodes are restructured. Typically, leaf-level nodes with the same occupancy are merged and promoted to a higher-level node.
Currently, this step is executed on the CPU by sending the voxel data back to the CPU. However, this is also possible on GPUs as demonstrated by previous works in Sec.~\ref{sec:gpurecon}.
The pseudo-code for our whole ray shooting procedure is given in Algorithm \ref{alg:GPU_rayTracing}.

%We upload the information to GPU such as sensor position, each ray direction, and a set of AABBs, and return all intersected voxels with the ray including occupancy state.
%All ray shooting is performed by the GPU in parallel, and when the ray traverses BVH and intersects AABB, it checks whether AABB is a node with the lowest-level depth of the octree, and if not, subdivides it.
%Then, the voxel that the ray passes through is set as a free state, and the voxel that includes the end point is set as an occupied state.

\section{EXPERIMENTS AND RESULTS}

In this section, we show our experimental results for GPU-based map building and also compare them against a state-of-the-art implementation.

\subsection{Evaluation of Ray Shooting}

\begin{figure*}[htb]
    \centering
     \begin{subfigure}[b]{0.16\linewidth}
        \centering
         \includegraphics[width=\linewidth]{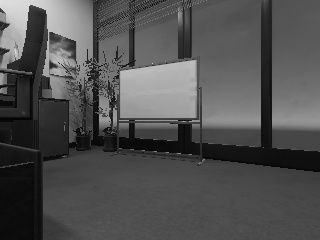}
     \end{subfigure}
    %  \hfill
     %\hspace{0.01em}
     \begin{subfigure}[b]{0.16\linewidth}
         \centering
         \includegraphics[width=\linewidth]{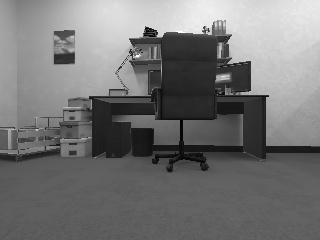}
     \end{subfigure}
    %  \hfill
    %\hspace{0.01em}
     \begin{subfigure}[b]{0.16\linewidth}
         \centering
         \includegraphics[width=\linewidth]{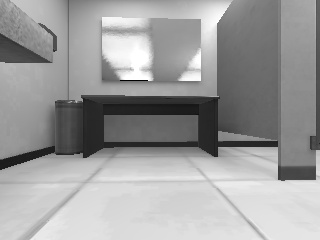}
     \end{subfigure}
    %  \hfill
    %\hspace{0.01em}
     \begin{subfigure}[b]{0.16\linewidth}
         \centering
         \includegraphics[width=\linewidth]{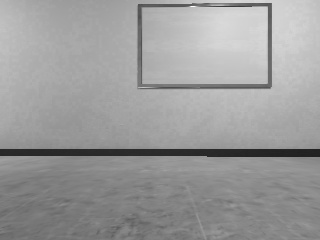}
     \end{subfigure}
      %  \hfill
    %\hspace{0.01em}
     \begin{subfigure}[b]{0.16\linewidth}
         \centering
         \includegraphics[width=\linewidth]{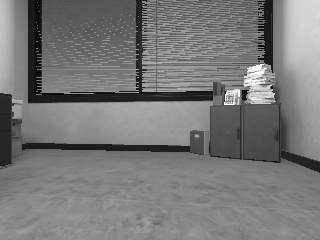}
     \end{subfigure}
     %  \hfill
    %\hspace{0.01em}
     \begin{subfigure}[b]{0.16\linewidth}
         \centering
         \includegraphics[width=\linewidth]{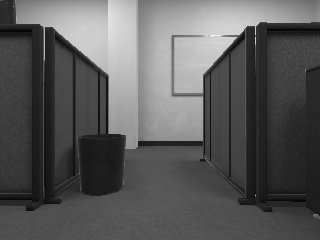}
     \end{subfigure}
     %%%
     %\hspace{0.01em}
     \begin{subfigure}[b]{0.16\linewidth}
        \centering
         \includegraphics[width=\linewidth]{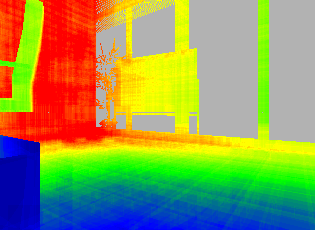}
         \caption{\label{fig:a}}
     \end{subfigure}
    %  \hfill
     %\hspace{0.01em}
     \begin{subfigure}[b]{0.16\linewidth}
         \centering
         \includegraphics[width=\linewidth]{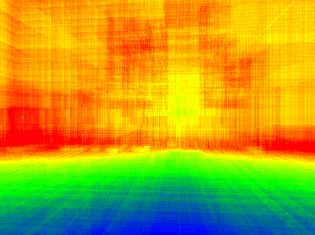}
         \caption{\label{fig:b}}
     \end{subfigure}
    %  \hfill
    %\hspace{0.01em}
     \begin{subfigure}[b]{0.16\linewidth}
         \centering
         \includegraphics[width=\linewidth]{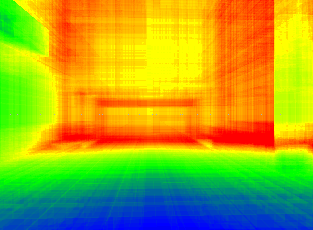}
         \caption{\label{fig:c}}
     \end{subfigure}
    %  \hfill
    %\hspace{0.01em}
     \begin{subfigure}[b]{0.16\linewidth}
         \centering
         \includegraphics[width=\linewidth]{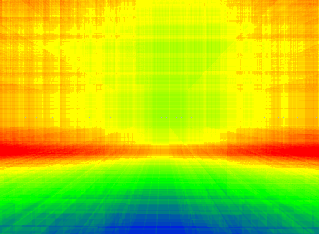}
         \caption{\label{fig:d}}
     \end{subfigure}
      %  \hfill
    %\hspace{0.01em}
     \begin{subfigure}[b]{0.16\linewidth}
         \centering
         \includegraphics[width=\linewidth]{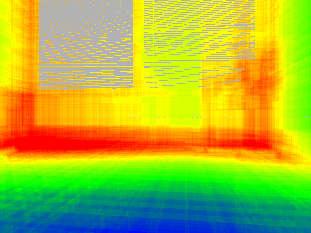}
         \caption{\label{fig:e}}
     \end{subfigure}
     %  \hfill
    %\hspace{0.01em}
     \begin{subfigure}[b]{0.16\linewidth}
         \centering
         \includegraphics[width=\linewidth]{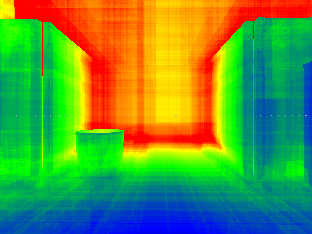}
         \caption{\label{fig:f}}
     \end{subfigure}
    \caption{\label{fig:views}
     Results of GPU-based ray shooting from different view points navigating inside a complex virtual building. The top row is the target scene. The bottom row is the corresponding hit count of rays with voxels in the space; as the color changes from blue to red, more voxels are intersected with rays; gray indicates that the distance sensor did not obtain point cloud due to reflections in the environment. }
\end{figure*}

\begin{table*}[]
\caption{Comparisons of Ray Shooting Performance on Octomap (CPU) and Ours (GPU), and Timing Breakdown}
\centering
\begin{tabular}{clccccccc}
\hline
\multicolumn{2}{c}{Benchmarking Views}                              & (a) & (b) & (c) & (d) & (e) & (f) \\ \hline
\multicolumn{2}{c}{\# of Free, Occupied or Unknown Voxels in Octree}                  & 25,515 & 33,806 & 34,412 & 21,832 & 23,609 & 31,215    \\ \hline
Octomap (CPU)                  & Ray Shooting (ms)        & 886.83 & 1,446.52 & 1,349.87 & 1,294.31 & 1,211.58 & 1,236.16    \\ \hline
\multirow{3}{*}{Ours (GPU)} & Build BVH (ms)           & 0.59 & 0.63 & 0.58 & 0.55 & 0.54 & 0.68    \\
% \cline{2-8} 
                           & Ray Shooting (ms)        & 1.42 & 1.67 & 2.07 & 1.14 & 1.11 & 1.59    \\ 
                        %   \cline{2-8} 
                           & Readback from GPU to CPU (ms) & 14.19 & 14.63 & 14.39 & 14.23 & 14.33 & 14.29     \\ \hline
\end{tabular}
\label{tab:perf}
\end{table*}

\begin{figure}[htb]
     \centering
      \begin{subfigure}[b]{0.93\linewidth}
          \centering
         \includegraphics[width=\linewidth]{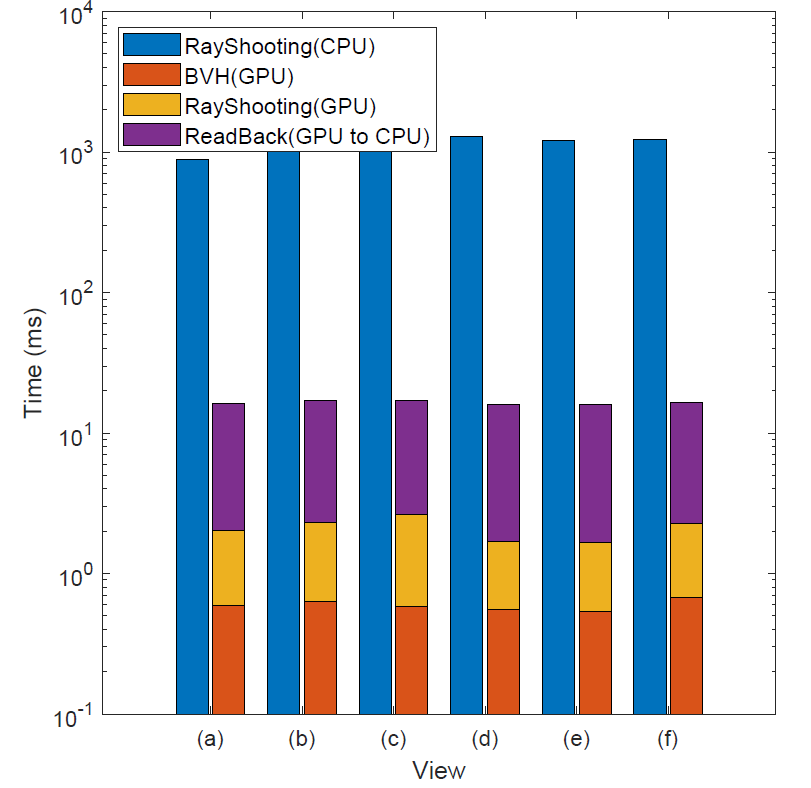}
        
      \end{subfigure}
     \hfill
    
    \caption{\label{fig:time_res}
        Relative Performance Comparisons of Ray Shooting between Octomap (CPU in blue bars) and Ours (GPU in gold bars) in the Logarithmic Scale. BVH construction and readback from GPU to CPU are denoted in orange and purple bars.
        }
\end{figure}

GPU-based ray shooting for map building was implemented on a 64bit Windows 10 operating system and Microsoft Visual Studio 2017 C++ with AMD's Ryzen 7 3700X CPU, NVIDIA's RTX 2080 GPU, and 16GBs RAM. We used DirectX's DXR to drive GPU-accelerated ray tracing on RTX.
As a benchmarking platform, we employed a virtual indoor environment built-in Tesse-Unity simulator \cite{YadBat19} where a mobile robot equipped with a stereo camera navigated around this environment to collect a point cloud data set and build an octree-based map. Robot localization is assumed to be given and exact.  

In order to test the benchmarks, while driving through a virtual building, at a random view point, a point cloud is acquired from the scene using a range sensor. The acquired point cloud data is built into an octree with a maximum depth of 16, corresponding to 25K$\sim$34K leaf-level voxels. The ray-shooting step, which is the main bottleneck during map building, is implemented on GPU and the rest of the building steps is done on CPU using Octomap.  
Figure~\ref{fig:views} illustrates six views from different viewpoints.
Here, we shoot 320$\times$240 rays (76,800 rays) per each view to collect and identify octree cells. 
We measure the ray shooting time using the dispatchRay function of RTX, which queries the elapsed time of ray tracing performed on the GPU. 
%For all rays, the execution time was spent in searching for occupied and free voxels hitting the ray.

For the sake of comparison, ray-shooting time on CPU using Octomap was also measured with the same experimental condition with the same camera position, direction, and ray length using the same point cloud and the same number of voxels of the octree. The CPU implementation was measured on Ubuntu 20.04 with Intel Core i7-7700HQ CPU @ 2.8GHz and 16 GBs of RAM.

Table~\ref{tab:perf} is the result of evaluating ray shooting time for each view shown in Fig.~\ref{fig:views}. In the case of GPU-based ray shooting, we further included the BVH construction time on the GPU and the time for reading back the intersected voxels from GPU to CPU. 
Fig.~\ref{fig:time_res} shows that GPU-based ray shooting can be performed three-orders-of-magnitude faster than CPU-based ray shooting on average excluding GPU-CPU readback. Even though GPU-CPU readback time is included, the performance improvement is still two-orders-of-magnitude faster than the CPU version, and the readback time may be removed by updating the octree and occupancy on the GPU using a technique like \cite{liu2014exact,kim2018dynamic}.
%Although GPU-CPU readback is a costly transfer operation, in the case of GPU-based ray shooting including readback time, it is almost 72 times faster than CPU-based ray shooting.    

%Fig.~\ref{fig:time_res} illustrates the contents of Table~\ref{tab:perf}.

\section{CONCLUSIONS}

In this paper, we propose GPU-based ray shooting to improve the ray shooting performance in the state-of-the-art PVM algorithm such as Octomap. The main idea is based on the use of recent ray-tracing RTX GPU and map voxel grids of an octree to a set of AABBs and employ massively parallel ray shooting on them using the GPUs to find free and occupied voxels. The observed speedup in ray-shooting is significant. There are limitations to our current work. First, the octree itself is not maintained on GPUs but CPU. Thus, it is required that the newly found voxels need to be readback from GPUs to CPU, which takes almost 10X more than the ray shooting itself. However, this can be addressed in the future using various GPU-based octree maintenance techniques as explained in Sec.~\ref{sec:gpurecon}.

%A conclusion section is not required. 

%\addtolength{\textheight}{-12cm}   % This command serves to balance the column lengths
                                  % on the last page of the document manually. It shortens
                                  % the textheight of the last page by a suitable amount.
                                  % This command does not take effect until the next page
                                  % so it should come on the page before the last. Make
                                  % sure that you do not shorten the textheight too much.

%%%%%%%%%%%%%%%%%%%%%%%%%%%%%%%%%%%%%%%%%%%%%%%%%%%%%%%%%%%%%%%%%%%%%%%%%%%%%%%%

%%%%%%%%%%%%%%%%%%%%%%%%%%%%%%%%%%%%%%%%%%%%%%%%%%%%%%%%%%%%%%%%%%%%%%%%%%%%%%%%

%%%%%%%%%%%%%%%%%%%%%%%%%%%%%%%%%%%%%%%%%%%%%%%%%%%%%%%%%%%%%%%%%%%%%%%%%%%%%%%%
%\section*{APPENDIX}

%Appendixes should appear before the acknowledgment.

\section*{ACKNOWLEDGMENT}

This project was supported in part by the ITRC/IITP program (IITP-2020-0-01460) and the NRF (2017R1A2B3012701) in South Korea.

%%%%%%%%%%%%%%%%%%%%%%%%%%%%%%%%%%%%%%%%%%%%%%%%%%%%%%%%%%%%%%%%%%%%%%%%%%%%%%%%

%References are important to the reader; therefore, each citation must be complete and correct. If at all possible, references should be commonly available publications.

% \begin{thebibliography}{99}

% % check out the format
% \bibitem{c1}  Nvidia,   “Nvidia   turing   gpu   architecture,”   2018.   [On-line].    Available:    https://www.nvidia.com/content/dam/en-zz/Solutions/design-visualization/technologies/turing-architecture/NVIDIA-Turing-Architecture-Whitepaper.pdf

% \bibitem{c2} M.Stich,“Introduction to NVIDIA RTX and DirectX Ray Tracing,” 2018. [Online]. Available:https://devblogs.nvidia.com/introduction-nvidia-rtx-directx-ray-tracing/

% \end{thebibliography}

\bibliographystyle{IEEEtran}
\bibliography{reference}

\end{document}